\documentclass{article}

\usepackage{PRIMEarxiv}

\usepackage[utf8]{inputenc} 
\usepackage[T1]{fontenc}    
\usepackage{hyperref}       
\usepackage{url}            
\usepackage{booktabs}       
\usepackage{amsfonts}       
\usepackage{nicefrac}       
\usepackage{microtype}      
\usepackage{lipsum}
\usepackage{fancyhdr}       
\usepackage{graphicx}       
\usepackage{xcolor}         
\usepackage{multirow}
\usepackage{caption}
\usepackage{amsmath}
\usepackage{subcaption}

\graphicspath{{figures/}}     

\title{Learnable Visual Words for Interpretable Image Recognition}

\author{%
  Wenxiao Xiao \\ 
  Department of Computer Science\\
  Brandeis University\\
  \texttt{wenxiaoxiao@brandeis.edu} \\
  \And
  Zhengming Ding \\
  Department of Computer Science \\
  Tulane University \\
  \texttt{zding1@tulane.edu} \\
  \And
  Hongfu Liu \\
  Department of Computer Science\\
  Brandeis University\\
  \texttt{hongfuliu@brandeis.edu} \\
}

\begin{document}
\maketitle

\begin{abstract}
   To interpret deep models' predictions, attention-based visual cues are widely used in addressing \textit{why} deep models make such predictions. Beyond that, the current research community becomes more interested in reasoning \textit{how} deep models make predictions, where some prototype-based methods employ interpretable representations with their corresponding visual cues to reveal the black-box mechanism of deep model behaviors. However, these pioneering attempts only either learn the category-specific prototypes and deteriorate their generalizing capacities, or demonstrate several illustrative examples without a quantitative evaluation of visual-based interpretability with further limitations on their practical usages. In this paper, we revisit the concept of visual words and propose the Learnable Visual Words (LVW) to interpret the model prediction behaviors with two novel modules: semantic visual words learning and dual fidelity preservation. The semantic visual words learning relaxes the category-specific constraint, enabling the general visual words shared across different categories. Beyond employing the visual words for prediction to align visual words with the base model, our dual fidelity preservation also includes the attention guided semantic alignment that encourages the learned visual words to focus on the same conceptual regions for prediction. Experiments on six visual benchmarks demonstrate the superior effectiveness of our proposed LVW in both accuracy and model interpretation over the state-of-the-art methods. Moreover, we elaborate on various in-depth analyses to further explore the learned visual words and the generalizability of our method for unseen categories.
\end{abstract}


\section{Introduction}
Model interpretation aims to explain the black-box base model in a semantically understandable way and preserve high fidelity with model outputs~\cite{li2021interpretable,molnar2020interpretable,zhang2021survey}. Although interpretative models are not designed to pursue higher performance than the base model, they are of great importance, especially in the deep learning era. With a clear understanding, model designers can diagnose errors and potential biases embedded in deep models; model users are confident and relaxed to rely on model predictions. For tablet data, the original features with physical meanings are used to interpret the model prediction. LIME~\cite{ribeiro2016should} and SHAP~\cite{lundberg2017unified} are two representative explainable models, which learn linear models to locally fit the base model's outputs with sample perturbations. For visual data, attention-based visual cues are widely adopted in interpreting deep models' predictions~\cite{correia2021attention}. Grad-Cam~\cite{selvaraju2017grad,selvaraju2020grad} is a popular technique for visualizing the region where a convolutional neural network model looks at. It uses the category-specific gradient information flowing into the final convolutional layer of a CNN to produce a coarse localization attention map of the important regions/pixels in the image. Following this direction, several studies further employ weakly supervised information directly on attention maps during the training stage to reduce the attention bias~\cite{Li2018TellMW, srinivas2019full, chaudhari2021attentive}.

Beyond the above studies addressing \textit{why} deep models make their predictions at the generally coarse level, some pioneering attempts have been taken to reason \textit{how} deep models make predictions at the fine-grained level. ProtoPNet~\cite{chen2019looks} learns a predetermined number of prototypes per category and proposes a prototypical part network with a hidden layer of prototypes representing the activated patterns. By learning these prototype-based interpretable representations, ProtoPNet explains the model reasoning process by dissecting the query image into several prototypical parts and interpreting these prototypes with training images in the same category. Later, ProtoTree~\cite{nauta2021neural} and ProtoPShare~\cite{rymarczyk2021protopshare} extend ProtoPNet by decision trees and prototype pruning to achieve global interpretation and reduce the model complexity, respectively. Unfortunately, these studies either learn the category-specific prototypes that deteriorate their generalizing capacities, or only demonstrate several illustrative examples without a quantitative evaluation of visual-based interpretability (Figure~\ref{fig:seen} demonstrates their discovered meaningless or irrelevant prototypes despite their high recognition accuracy). Moreover, they only evaluate on two object-cropped visual datasets, and leave their performance on other general visual datasets and samples from unseen categories on the shelf.

\textbf{Contributions}. The process of ProtoPNet recalls us of the bags-of-visual words~\cite{csurka2004visual,sivic2003video}, a popular image representation technique before the deep learning era. It treats an image as a document and visual words are defined by the keypoints/descriptors/patches that are used to construct vocabularies. Then the image can be represented as a histogram over the occurrences of these visual words. It is worthy to note that these visual words with semantic meanings across different images can also be used for model interpretation, which is similar to the prototypes in ProtoPNet. The major difference lies in that conventional visual words are usually hand-crafted, pre-defined and independent of the downstream learning tasks. In light of this, we propose the Learnable Visual Words (LVW) to overcome the aforementioned drawbacks of the prototype-based interpretative methods. Technically, our model consists of two modules, semantic visual words learning and dual fidelity preservation. The semantic visual words learning relaxes the category-specific constraint enabling the generic visual words shared across different categories, while the dual fidelity preservation encourages the learned visual words to behave similarly to the base model in both prediction and model attention. Our major contributions are summarized as follows:
\begin{itemize}
    \item In the semantic visual words learning, we relax its category-specific constraint and further simplify ProtoPNet to achieve cross-category visual words and increase the generalization of model interpretation, rather than adding any new terms.
    \item In the dual fidelity preservation, we encourage the learned visual words to preserve high fidelity with the base model in terms of both prediction and model attention. Additionally, we further design a measurement to quantitatively evaluate the visual-based interpretation.
    \item We demonstrate the superior effectiveness of our model on six visual benchmarks over the state-of-the-art prototype-based methods in both accuracy and visual-based interpretation and explore the generalization of our model by interpreting unseen categories. 
\end{itemize}

\textbf{Related Work}.
Research efforts in interpretable explanations of a Convolutional Neural Network (CNN) can be generally divided into \textit{posthoc} and \textit{self-interpretable} genres. Posthoc methods attempt to build an extra explainer for the pre-trained black-box model to interpret its prediction. Approaches including saliency visualization based on backpropagation \cite{springenberg2014striving, zhou2016learning, zhang2018top, bach2015pixel, sundararajan2017axiomatic, shrikumar2017learning, smilkov2017smoothgrad, fong2017interpretable, Rebuffi2020ThereAB} and activation maximization caused by perturbation \cite{simonyan2013deep, zeiler2014visualizing, petsiuk2018rise, fong2019understanding, Kapishnikov2019XRAIBA, dabkowski2017real, ancona2017towards} to identify influential parts for the black-box model's prediction. However, visualizing the salient areas does not explain \textbf{how} the black box makes such decisions. Other posthoc methods \cite{ghorbani2019towards, zhang2020invertible, olah2018building, akula2020cocox, yeh2020completeness, koh2020concept, kim2018interpretability, chen2020concept} obtain interpretable Concept Activation Vectors from pre-segmented feature maps and interpret the CNN model with these concepts. Alternatively, self-interpretable methods aim to directly learn explanatory representations during training, instead of disentangling the pre-trained black box. ProtoPNet~\cite{chen2019looks} introduces a case-based study that interprets the reasoning process of the black-box model with category-specific prototypes associated with image patches from training data, which answers how deep models make predictions by linking the test image with interpretable prototypes. Recent studies \cite{DBLP:journals/corr/abs-2112-02902, DBLP:journals/corr/abs-2111-15000} extend ProtoPNet in various directions. For instance, ProtoTree~\cite{nauta2021neural} combines prototype learning with decision trees, resulting in an interpretable decision path consisting of prototypes. Similarly, HPnet~\cite{hase2019interpretable} hierarchically organizes prototypes to classify objects at every level in a predefined class taxonomy. ProtoPShare\cite{rymarczyk2021protopshare} further groups similar prototypes in a pre-trained ProtoPNet model with a data-dependent merge-pruning strategy. Meanwhile, TesNet~\cite{Wang_2021_ICCV} extracts prototypes from class-specific embedding subspaces defined on the Grassmann Manifold.

\section{Method}

\label{method}

\subsection{Framework Overview}
\begin{figure*}
  \begin{minipage}[c]{0.7\textwidth}
    \includegraphics[height=5.3cm, width=11.3cm]{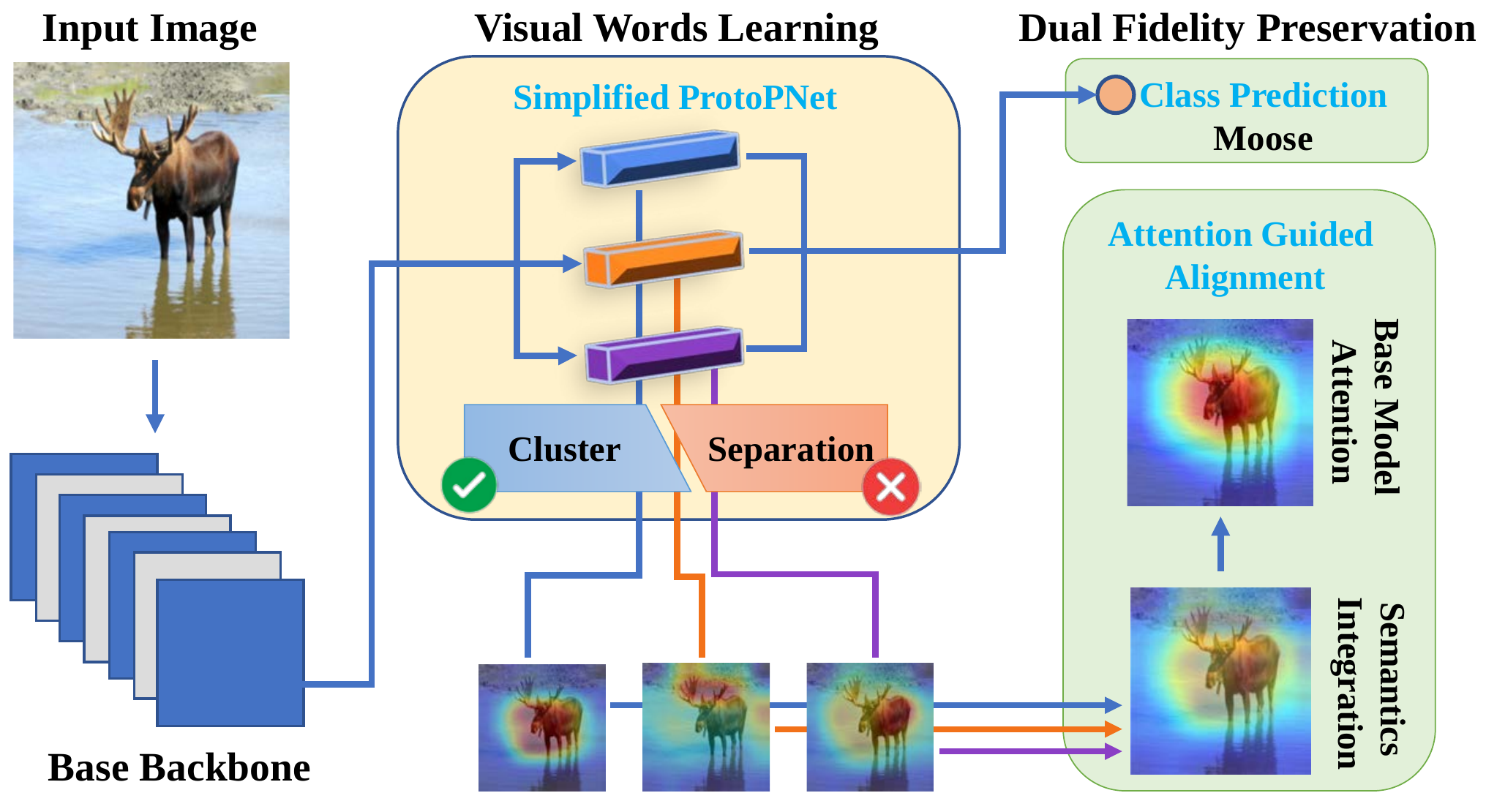}
  \end{minipage}\vspace{-3mm}\hfill
  \begin{minipage}[c]{0.3\textwidth}\vspace{3mm}
    \caption{Illustration of our proposed Learnable Visual Words (LVW) model for image recognition interpretation. Visual Words Learning module simplifies ProtoPNet and learns cross-category semantic visual words to describe the samples; the Dual Fidelity Preservation module enhances the learned visual words to preserve high fidelity with the base model in terms of both prediction and attention.} \label{fig:framework}
  \end{minipage}\vspace{-2mm}
\end{figure*}

We introduce the framework of our Learnable Visual Words (LVW) model to extract meaningful semantic information for interpreting the base CNN models in Figure~\ref{fig:framework}, which consists of visual words learning and dual fidelity preservation. The visual words learning module aims to extract semantics shared across all categories, while the dual fidelity preservation module guides these learned visual words to focus on the regions of the image that base CNN models attend to when making predictions and also preserves the similar predictive ability of the base model. In order to learn cross-class visual words, our semantic learning module simplifies ProtoPNet~\cite{chen2019looks}, by removing its separation loss, which is used for finding category-specific prototypes for each category. We keep ProtoPNet's clustering loss so that the learned visual words are closely related to the training images, and at the same time, contain various shared semantic information from all classes. The dual fidelity preservation module assures the learned visual words preserve high fidelity with the base model in terms of both prediction and attention in a similarity-based approach. For each image sample, the similarity scores between one visual word and all patches of the sample image's backbone convolution output are calculated, resulting in a similarity heatmap that represents how strong this visual word matches different parts of the image, the same as ProtoPNet. The maximum value of the similarity heatmap for each visual word is then used for making the final prediction by the fully connected layer with cross-entropy loss. To achieve interpretability for the prediction made by the CNN backbone, the attention guided alignment module combines the similarity heatmaps of the sample image's top $k$ visual words with max-pooling, and guides the pooled heatmap with the base CNN model's class-level attention. With attention alignment, the learned visual words focus on the same areas used by the base model for making predictions, preserving the attention of the base model. 

\subsection{Learnable Visual Words for Interpretable Model}

Since our model simplifies ProtoPNet in visual words learning, we first introduce the base backbone model and ProtoPNet to provide enough preliminaries, and then elaborate on our innovations.

\textbf{Preliminaries}.\label{Prelim} Let the base backbone model (e.g., ResNet~\cite{he2016deep}, VGG~\cite{simonyan2014very}) be trained in a supervised setting with $N$ labeled training samples $\mathcal{X}=\{(x_i, y_i)\}, 1$$\leq$$i$$\leq$$N$ associated with $C$ different categories. ProtoPNet~\cite{chen2019looks}, a prototype-based explanation method for a CNN model, consists of a convolutional layer $f$, a prototype layer $g$ and a prediction layer $h$. It first extracts the feature map $Z_i = f(x_i)$ with a dimension of $H$$\times$$W$$\times$$D$ from the last convolutional layer's output of the base model. Then ProtoPNet learns total $M$ prototypes or visual words $\mathcal{V}=\{v_j\}|_{j=1}^M$ with the shape of $H'$$\times$$W'$$\times$$D$, $H'$$<$$H$ and $W'$$<$$W$. Their prototype layer $g$ computes an activation map consisting of the inverted $L_2$-distance between one learned visual word $v_j$ and all $H'$$\times$$W'$$\times$$D$-shape patches of $Z$. Then the activation map is resized back to the original image size, resulting in an activation heatmap $A_j$ to represent how strong $v_j$ matches each part of the input image. ProtoPNet takes the maximum value in $A_j$, i.e., $\mathrm{max}(A_j)$ as a similarity score between visual word $v_j$ and the input image. Later, a fully-connected layer $h$ makes the prediction based on similarity scores $g \circ f(x_i)=[\mathrm{max}(A_1),...,\mathrm{max}(A_M)]$ associated with all $M$ learned visual words. 

\textbf{Visual Words Learning}. Our visual words learning module is a simplified version of ProtoPNet~\cite{chen2019looks} where we keep its clustering loss but discard the separation loss. The clustering loss encourages all training images to be close to some learned visual words; on the other hand, removing the separation loss relaxes the category-specific constraint and allows the learned visual words to be shared across categories. Mathematically, the cluster loss is defined as:
\begin{equation} \label{eq:cluster}
\mathcal{L}_{c} = \frac{1}{N}\sum_{i=1}^{N}\underset{ v_j}{\operatorname{\min}}\underset{z \in Z_i}{\operatorname{\min}}||z - v_j||^2,
\end{equation}
where $z$ is a single patch of shape $H'$$\times$$W'$$\times$$D$ in $Z_i$.

\textbf{Attention Guided Semantic Alignment}. Beyond employing the visual words for prediction, our dual fidelity preservation also includes the attention guided semantic alignment, which requests the learned visual words to focus on the same region as the base CNN model. To achieve this goal of attention fidelity, the alignment module combines the activation heatmaps of the top $k$ visual words with the highest similarity after max-pooling, resulting in a single activation heatmap $A^{k}$ that represents how strong these $k$ visual words are present in the input image $x$. The attention alignment abides by the loss term as follows:
\begin{equation} \label{eq:alignment}
\mathcal{L}_{a} = \frac{1}{N}\sum_{i=1}^{N}||\bar{A}_i - A^{k}_i||^2,
\end{equation}
where $\bar{A}_i$ is the input class-level attention of the base CNN model on image $x_i$, and $A^{k}_i$ is the combined activation heatmap pooled from the $k$ top visual words that are semantically closest to $x_i$.

\subsection{Training Protocol and Objective Functions}
For training, we adopt the same three-step procedure as ProtoPNet\cite{chen2019looks}: (1) optimization before the last layer; (2) projection of visual words; (3) optimization of the last layer.

\textbf{Optimization Before the Last Layer}. In this stage, we fix the fully connected layer $h$ and train the convolutional layers $f$ and visual words layer $g$. For each category, the connections between $M/c$ visual words are set to 1 and the rest connections are set to -0.5, encouraging the model to learn some visual words that are closely related to this category. Different from ProtoPNet~\cite{chen2019looks}, we remove the separation loss, so that the learning module does not force the learned visual words to be category-specific. The overall objective function for this stage is:
\begin{equation} \label{eq:alignment}
\underset{f,g}{\operatorname{\min}}\frac{1}{N}\sum_{i=1}^{N}\mathcal{L}_{cls}(h \circ g \circ f(x_i)) + \alpha\mathcal{L}_{c} + \beta\mathcal{L}_{a},
\end{equation}
where $g \circ f(x_i)$ is a vector containing similarity scores between $M$ visual words to $x_i$, $\alpha$ and $\beta$ are two trade-off parameters.

\textbf{Projection of Visual Words}. During this stage, we project each visual word $v_j$ onto the nearest patch across all training images to further ensure that the learned visual words contain various semantic information from the training data. The project can be described as follows:
\begin{equation} \label{eq:projection}
v_j \Leftarrow \space \underset{z \in \mathcal{Z}}{\operatorname{\min}}||z-v_j||^2,
\end{equation}
where $\mathcal{Z}$ is the collection of all latent patches of shape $H'$$\times$$W'$$\times$$D$ contained in $\mathcal{X}$.

\textbf{Optimization of the Last Layer}. In this stage, we only optimize the fully connected layer $h$ with fixed convolutional layers $f$ and visual words layer $g$. This training stage uncovers the semantic associations of each learned visual word across all categories. Following ProtoPNet, we also add $L_1$-regularization during this training stage and the objective function is:
\begin{equation} \label{eq:last}
\underset{h}{\operatorname{\min}}\frac{1}{N}\sum_{i=1}^{N}\mathcal{L}_{cls}(h \circ g \circ f(x_i)) + \gamma\sum_{w \in W_h}|w|,
\end{equation}
where $W_h$ is the collection of all weights in the fully connected layer and $\gamma$ is the trade-off parameter.

\section{Experiments}
We demonstrate the performance of our learnable visual words in two aspects: classification ability and attention-based model interpretability. We first introduce the experimental setup, then report the algorithmic performance with extended examples, and finally provide various in-depth analyses.

\subsection{Experimental Setup}
\textbf{Datasets}.\label{datasets} We include six datasets for performance evaluation. (1) \textit{STL10}~\cite{pmlr-v15-coates11a} is a popular image classification benchmark containing 13,000 labeled images from 10 object classes. 5,000 images are used for training while the remaining images are kept for testing. (2) \textit{Oxford 102 Flower}~\cite{Nilsback08} is a fine-grained image classification dataset including 102 flower categories. Each class consists of between 40 and 258 images. Based on its original split, we use its training set for training and its validation set for testing. (3) \textit{Oxford-IIIT Pets}~\cite{parkhi12a} has 37 categories including various breeds of dogs and cats, with roughly 200 images for each class and it is often used for fine-grained image classification tasks. (4) \textit{Stanford Dogs}~\cite{KhoslaYaoJayadevaprakashFeiFei_FGVC2011} contains 20,580 images with fine-grained annotation for 120 different dog breeds. The sample images are divided into 12,000 images for training and 8,580 images for testing. (5) \textit{Food-101} dataset~\cite{bossard14} is a large-scale fine-grained classification benchmark, which contains 101 food categories with 750 training and 250 test images per category. (6) \textit{AwA2}~\cite{8413121} contains 37,322 images in 50 animal categories. We choose this dataset specifically for checking if our method generalized well for non-fine-grained datasets. Notice that all sample images are not cropped, even for those datasets that come with Region of Interest annotation. 

\textbf{Baseline Models and Implementation Details}.\label{implementation} We compare our model to three prototype-based methods,\footnote{We notice two new variants of ProtoPNet, ProtoPool~\cite{DBLP:journals/corr/abs-2112-02902} and Deformable ProtoPNet~\cite{DBLP:journals/corr/abs-2111-15000}. We do not include these two methods for comparison as their papers are not formally published and their codes are still unavailable.} including ProtoPNet~\cite{chen2019looks}, ProtoTree~\cite{nauta2021neural}, and ProtoPShare~\cite{rymarczyk2021protopshare}. We choose {ResNet-34}~\cite{he2016deep} as the base CNN backbone for all three baseline models as well as our model. For ProtoPNet and ProtoPShare, 10 prototypes are selected per category. For ProtoTree, we train a tree of depth 10 so that each sample image will be associated with 10 prototypes in the test stage. For our method, we choose $M$$=$$5$$\times$$C$ visual words, where $C$ is the number of classes.

We implement our model in {PyTorch}~\cite{paszke2019pytorch} using one NVIDIA Titan V GPU.\footnote{Our code is available at \href{https://github.com/LearnableVW/Learnable-Visual-Words}{\textit{https://github.com/LearnableVW/Learnable-Visual-Words}}.} For all datasets, we fine-tune the {ResNet-34} base model pre-trained on \textit{ImageNet}~\cite{5206848} for each dataset, and then Grad-Cam~\cite{selvaraju2017grad,selvaraju2020grad} attention map is extracted with the fine-tuned {ResNet}. All the images are resize to 224$\times$224$\times$3 except for \textit{STL10}. We keep the original image size for \textit{STL10} as all its images have the same size. Random horizontal flip is applied during training as the only data augmentation procedure. The shape of convolutional output $f(x)$ is $H$$=$$W$$=$$7$ with $D$$=$$128$ channels. The shape of each visual word is set to $H'$$=$$W'$$=$$1$. The learning rate of the backbone feature extractor is set as 0.0001, while the learning rate the other parts is set as 0.003. $\alpha$, $\beta$ and $\gamma$ are set to 0.8, 10 and 0.0001, respectively. The number of visual words $k$ is set to 10 empirically for attention guided alignment training. All datasets are trained for 200 epochs, and we project visual words after every 10 epochs of training. We use ReLU as the activation function except for the last fully connected layer, in which we use the sigmoid function.

\textbf{Evaluation}. Recognition accuracy is used to evaluate the predictive ability of each model. In addition to accuracy, we also propose an Intersection over Union (IoU) coverage metric as an objective measurement for the model's interpretability as follows:
\begin{equation} \label{eq:iou}
\mathrm{IoU}_{q,m}=\frac{\mathcal{M}_{q}(\bar{A}) \cap \mathcal{M}_{q}(A^k)}{\mathcal{M}_{q}(\bar{A}) \cup \mathcal{M}_{q}(A^k)},
\end{equation}
where $\bar{A}$ is the class-level attention from the base model, and $A^k$ is the combined attention acquired from the $m$ most related visual words. Here, $\mathcal{M}_q$ means the binary masking function that masks out all the values that smaller than the $q$-th quantile of the attention map as 0 and keeps the rest values as 1. In our evaluation, we set $k=10$ and $q=50$ by default. Essentially, the designed IoU metric represents the amount of overlap between each sample's base model attention and the top $k$ visual words' attention. Thus, we use the average IoU across all test samples to measure the model's ability in explaining the decision made by the base CNN model.

\textbf{Visualization Protocol}. We adopt a similar protocol of visualizing learned visual words as ProtoPNet~\cite{chen2019looks}. For visual word $v_j$ globally, we identify the training image $x_i$ whose convolutional output $Z_i$ has the highest activation of $v_j$. The global visualization of $v_j$ is generated by overlaying the up-sampled activation heatmap $A_{j}$ on $x_i$. For any test image $x'$ locally, the activation heatmap $A_j$ of visual word $v_j$ on $x'$ represents the presence of $v_j$ on $x'$.

\begin{table}
  \caption{Performance of different interpretative models by accuracy and IoU coverage on six datasets}
  \label{performace}
  \centering
  \resizebox{1.\textwidth}{!}{
  \begin{tabular}{l|cc|cc|cc|cc|cc|cc}
    \toprule
    \multirow{2}{*}{Method} &
    \multicolumn{2}{c}{ \textit{STL10}} & 
    \multicolumn{2}{|c}{ \textit{Oxford Flower}} &
    \multicolumn{2}{|c}{ \textit{Oxford-IIIT Pets}} & 
    \multicolumn{2}{|c}{\textit{Stanford-Dogs}} & 
    \multicolumn{2}{|c}{\textit{Food-101}} & 
    \multicolumn{2}{|c}{\textit{AWA2}}\\ 
    
    \cmidrule(r){2-3}
    \cmidrule(r){4-5}
    \cmidrule(r){6-7}
    \cmidrule(r){8-9}
    \cmidrule(r){10-11}
    \cmidrule(r){12-13}
     &Acc &IoU &Acc &IoU &Acc &IoU &Acc &IoU &Acc &IoU &Acc &IoU \\
    \midrule
    ResNet-34~\cite{he2016deep} &87.40 &1.00 &98.17 &1.00 &92.28 &1.00 &79.03 &1.00 &81.60 &1.00 &93.66 &1.00\\
    \midrule
    ProtoPNet~\cite{chen2019looks} &86.33 &0.2199 &86.33 &0.3081 &82.12 &0.2190 &69.27 &0.4337 &74.93 &0.4135 &89.18 &0.2294\\
    ProtoTree~\cite{nauta2021neural} &89.58 &0.3228 &30.44 &0.3418 &62.41 &0.5113 &41.00 &0.4267 &36.37 &0.4173 &86.54 &0.6197 \\
    ProtoPShare~\cite{rymarczyk2021protopshare} &87.04 &0.2722 &87.04 &0.2958 &74.70 &0.3467 &68.70 &0.4585 &72.37 &0.4826 &84.10 &0.2571\\
    \midrule
    Ours w/o Atte. Align. &{89.55} &0.2209    &{89.85} &{0.3722}    &{85.47} &0.2583    &{70.72} &{0.2748}    &{75.03} &{0.3128}    &\textbf{91.16} &{0.2536}    \\ 
    Ours &\textbf{89.75} &\textbf{0.6465} &\textbf{91.81} &\textbf{0.7967} &\textbf{86.20} &\textbf{0.8086} &\textbf{73.81} &\textbf{0.7932} &\textbf{76.33} &\textbf{0.6969} &{90.85} &\textbf{0.7781}\\
    \bottomrule
  \end{tabular}
  }
\end{table}

\begin{figure}[t]
  \centering
  \includegraphics[height=16cm, width=\textwidth]{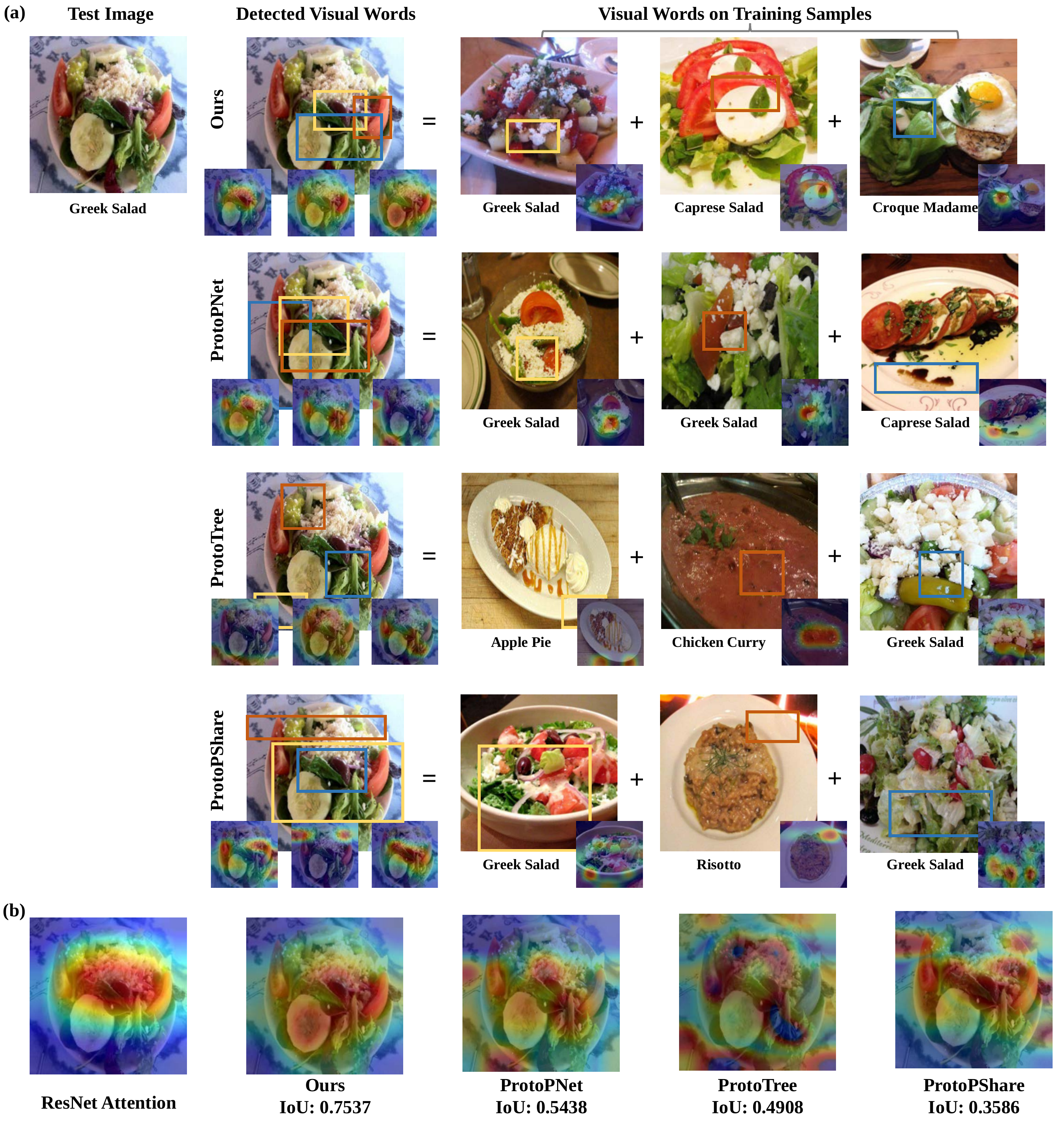}
  \caption{Visual demonstration of our model and other prototype-based methods for the same base model (ResNet-34) interpretation on \textit{Food-101}. (a) shows the detected visual words or prototypes on the test image and corresponding training samples, where the bounding boxes are generated according to the activated regions of visual words or prototypes. We use the same technique in ProtoPNet~\cite{chen2019looks} for bounding box generation. (b) shows the ResNet attention map and the integrated attention maps of visual words or prototypes with their IoU values by Eq.~\eqref{eq:iou}.}
  \label{fig:seen}\vspace{-5mm}
\end{figure}

\begin{figure}[t]
     \begin{subfigure}[b]{0.32\textwidth}
         \centering
         \includegraphics[height=4.5cm, width=\textwidth]{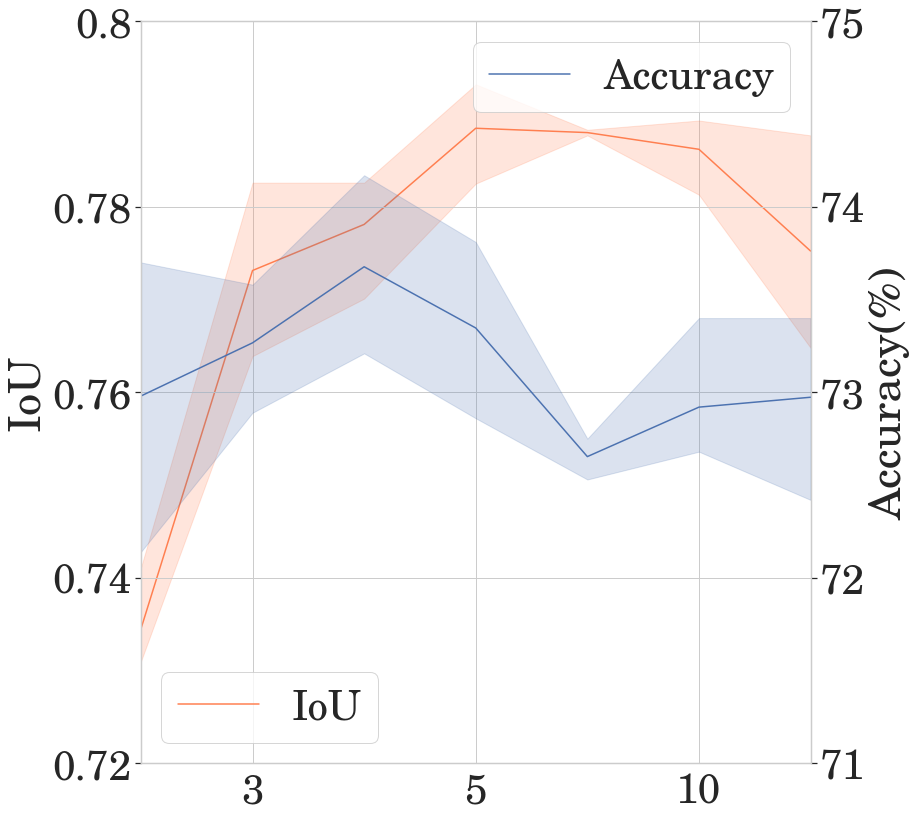}
         \caption{IoU \& Acc on \textit{Stanford-Dog}}
         \label{fig:k}
     \end{subfigure}
     \hfill
     \begin{subfigure}[b]{0.32\textwidth}
         \centering
         \includegraphics[height=4.5cm,width=\textwidth]{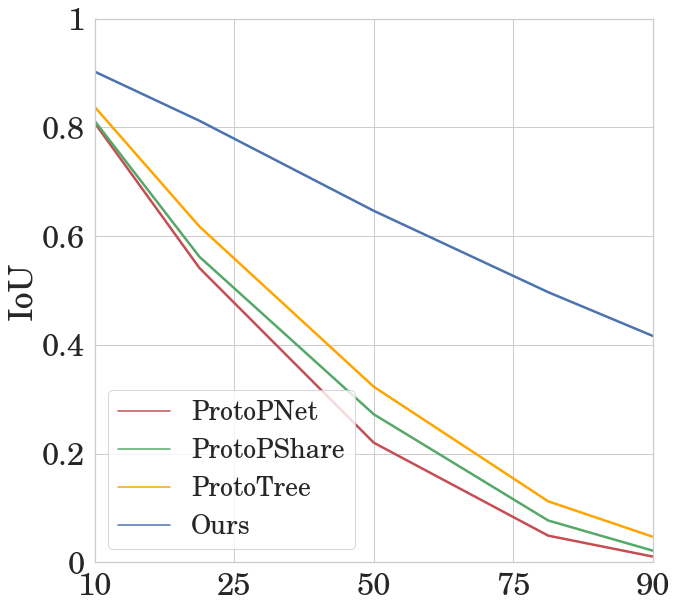}
         \caption{\textit{STL10}}
         \label{fig:stl10}
     \end{subfigure}
     \hfill
     \centering
     \begin{subfigure}[b]{0.32\textwidth}
         \centering
         \includegraphics[height=4.5cm,width=\textwidth]{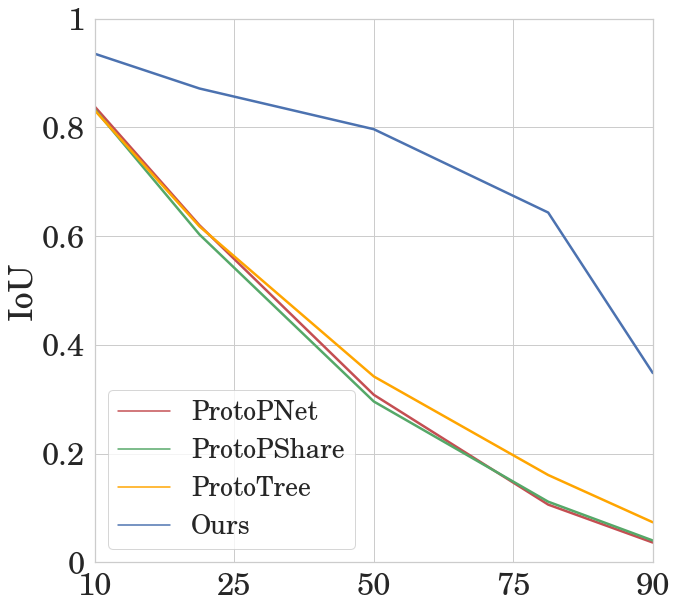}
         \caption{\textit{Oxford Flower}}
         \label{fig:flower}
     \end{subfigure}
     \hfill
     \begin{subfigure}[b]{0.32\textwidth}
         \centering
         \includegraphics[height=4.5cm,width=\textwidth]{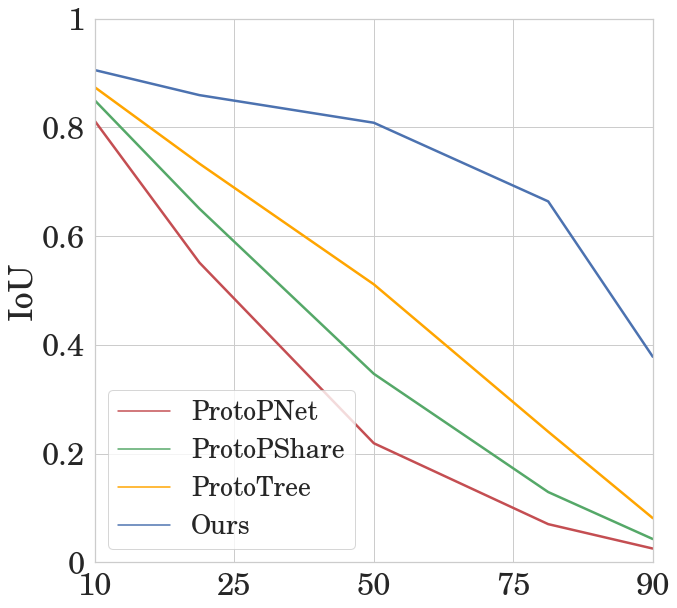}
         \caption{\textit{Oxford-IIIT Pets}}
         \label{fig:pets}
     \end{subfigure}
     \hfill
     \begin{subfigure}[b]{0.32\textwidth}
         \centering
         \includegraphics[height=4.5cm,width=\textwidth]{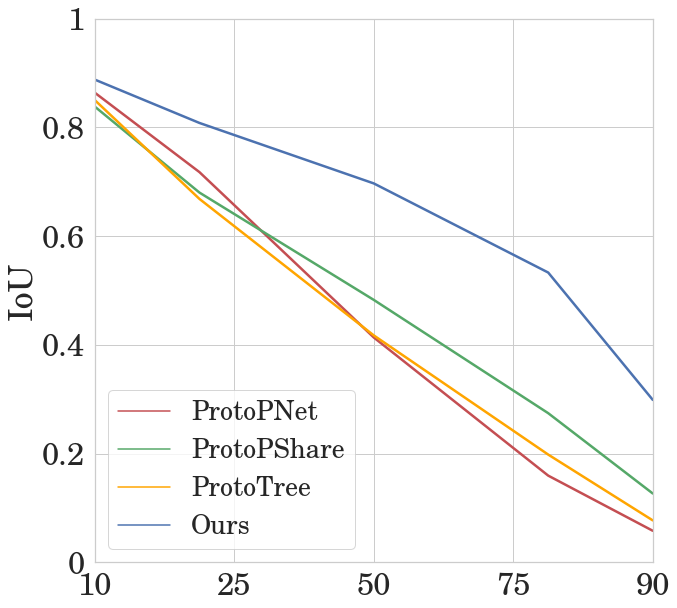}
         \caption{\textit{Food-101}}
         \label{fig:food}
     \end{subfigure}
     \hfill 
     \begin{subfigure}[b]{0.32\textwidth}
         \centering
         \includegraphics[height=4.5cm,width=\textwidth]{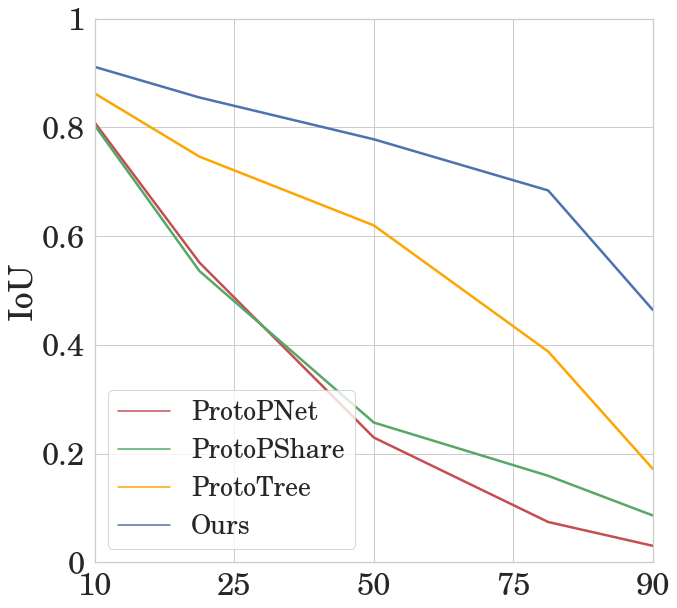}
         \caption{\textit{AWA2}}
         \label{fig:awa}
     \end{subfigure}
     \caption{(a) Performance of our method with different numbers of top visual words $k$. (b-f) IoU coverage for ours and baseline models with different quantile thresholds on \textit{STL10}, \textit{Oxford Flower}, \textit{Oxford-IIIT Pets}, \textit{Food-101} and \textit{AWA2}.}
    \label{fig:quatile}\vspace{-4mm}
\end{figure}

\subsection{Algorithmic Performance}
Here we comprehensively evaluate our learnable visual words model and three prototype-based interpretable recognition models with classification accuracy as well as the proposed IoU coverage metric. We also visually demonstrate a case study on \textit{Food-101} dataset. 

\textbf{Quantitative Evaluation}. Table~\ref{performace} shows the recognition accuracy and the proposed IoU coverage of all models on six datasets, in which we highlight the best results in bold. In addition to our complete model, we also report the performance of our model without attention guided alignment in order to check the efficacy of this module. In general, our proposed model achieves comparable recognition accuracy on these six datasets with the base ResNet-34 model. Our model sacrifices at most 6.36\% accuracy loss on \textit{Oxford Flower}~\cite{Nilsback08}, while, to our surprise, improving ResNet-34 by 2.35\% on \textit{STL10}~\cite{pmlr-v15-coates11a} dataset. Compared with the three baseline models, our full model attains the best recognition accuracy in 5 out of 6 datasets, only second to our model without attention alignment on \textit{AWA2}~\cite{8413121}, which boosts prediction accuracy over ProtoPNet~\cite{chen2019looks} only by removing the class-specific constraint. We assert that our model preserves, and possibly improves, the predictive ability of baseline ResNet-34. However, accuracy alone cannot properly estimate the visual-based interpretability of different models (We will provide more evidence in the next Visual Demonstration paragraph). Therefore, we also evaluate all methods with our proposed IoU coverage metric. Our model exceeds all baseline models by a large margin in terms of IoU coverage. At a minimum, our model outperforms baseline models by 0.1583 on \textit{AWA2}, while on datasets like \textit{Oxford Flowers} and \textit{STL10}~\cite{pmlr-v15-coates11a}, the IoU of our model can be twice as the IoU of the second-best baseline method, which suggests that our learned visual words preserve base model's attention and provide reasonable interpretation for the behavior of the base model. Notice that IoU coverage of the model without attention alignment is worse than ProtoPNet on datasets like \textit{Stanford-Dogs}~\cite{KhoslaYaoJayadevaprakashFeiFei_FGVC2011} and \textit{Food-101}\cite{bossard14} despite the improved accuracy, indicating that only relaxing category-specific constraint does not enhance its interpretability.

\textbf{Visual Demonstration}. We compare our model with the three prototype-based methods visually with a case-study example in Figure~\ref{fig:seen}. We choose one test image from \textit{Food-101}~\cite{bossard14} datasets and select three most related visual words or prototypes extracted by each method. We visualize them in the test images as well as their corresponding projected training images with the same protocol as ProtoPNet\cite{chen2019looks} in Figure~\ref{fig:seen}(a). Our method detects the visual words semantically corresponding to ``cheese crumbs,'' ``tomatoes,'' and ``green vegetables'' across three training categories. On the other hand, although the baseline models can also uncover prototypical patches from the sample images, the detected prototypes are more likely to focus on the meaningless or irrelevant semantics which provide little explanation for the prediction. For example, all three baselines identify prototypes related to the plate located on the edge of the test image. Those prototypes might help classify this test image as ``greek salad,'' but they are unable to interpret this decision. We conjure that without the guidance of base model attention, these prototype-based methods rely on those prototypes of the base model's attention area to make predictions, especially on the uncropped image data. In Figure~\ref{fig:seen}(b), both the IoU coverage values and the combined attention maps illustrate that the ResNet attention is well covered by our visual words, while the baseline methods focus more on the edge area of the test image. 
From this visual demonstration, we can see that our learnable visual words provide a more reasonable interpretation of how the base model makes predictions other than competitive methods.

\textbf{Hyperparameter Analysis}. We explore our model's sensitivity to  hyperparameters on different datasets. One key hyperparameter in our model is the number of top visual words $k$ used in attention guided semantic alignment. We choose different values of $k$ from 1 to 50 and conduct experiments on \textit{Stanford-Dogs} dataset. We train our model for each choice of $k$, and randomly select 3 different subsets of the original test set, each containing 50\% of total test images, to check our model robustness against different test data. As shown in Figure~\ref{fig:quatile}(a), prediction accuracy does not fluctuate for various choices of $k$. The IoU drops for smaller $k$ like 1 or 3, since we only encourage few visual words to focus on the class-level attention. For each $k$, the error between three random test sets is insignificant, considering our large IoU improvement. Moreover, the choice of quantile threshold $q$ in our designed measurement could also significantly affect the final results of IoU during the testing stage. To evaluate the robustness and generalizing ability, we plot the IoU coverage for 5 different quantile thresholds on the rest 5 datasets. Our model consistently outperforms the three baseline models on all benchmarks as shown in Figure~\ref{fig:quatile}(b-f). For $q=90$, the attention of baseline models barely overlaps with the base model attention, while our method still maintains at least 0.2 IoU coverage, which evinces our proposed method uncovers the most attended semantics in the base model.

\begin{figure}[t]
     \begin{subfigure}[b]{0.32\textwidth}
         \centering
         \includegraphics[height=4.5cm,width=\textwidth]{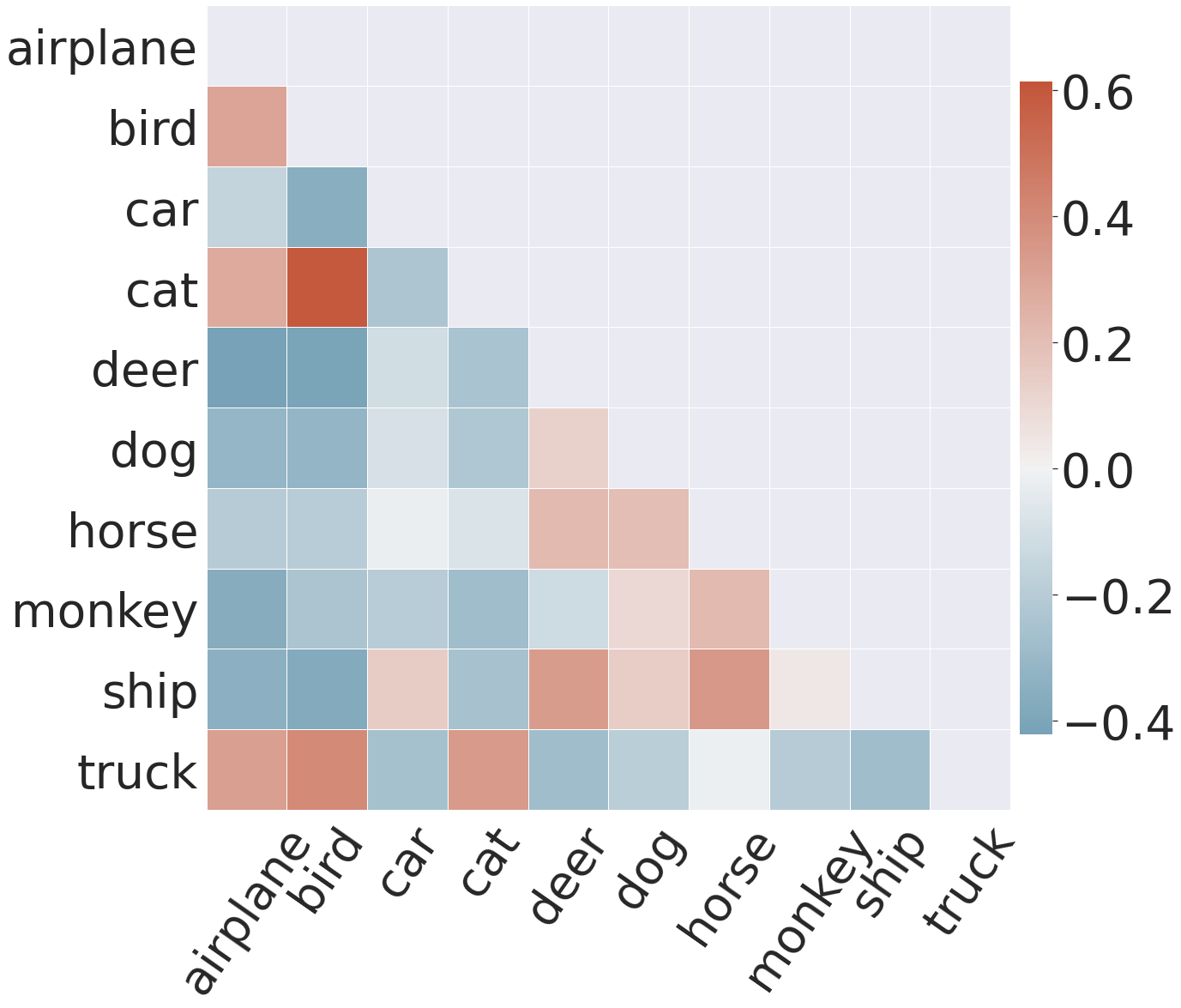}
         \captionsetup{justification=centering}
         \caption{Visual Words-Based\\ Category Similarity}
         \label{fig:htmp}
     \end{subfigure}
     \hfill
     \begin{subfigure}[b]{0.32\textwidth}
         \centering
         \includegraphics[height=4.5cm,width=\textwidth]{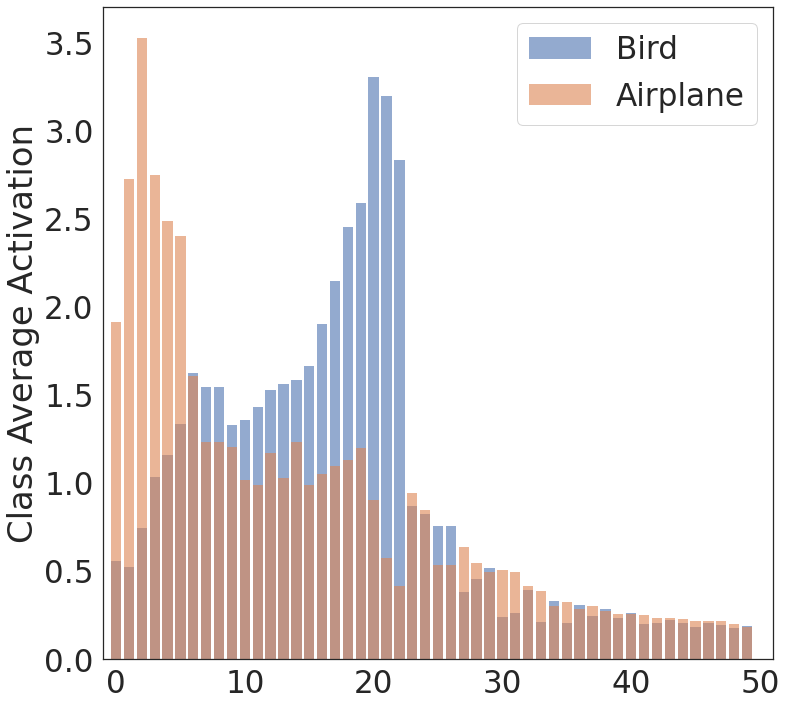}
         \captionsetup{justification=centering}
         \caption{Visual Words Activation\\ of Similar Categories}
         \label{fig:shared}
     \end{subfigure}
     \hfill
     \begin{subfigure}[b]{0.32\textwidth}
         \centering
         \includegraphics[height=4.5cm,width=\textwidth]{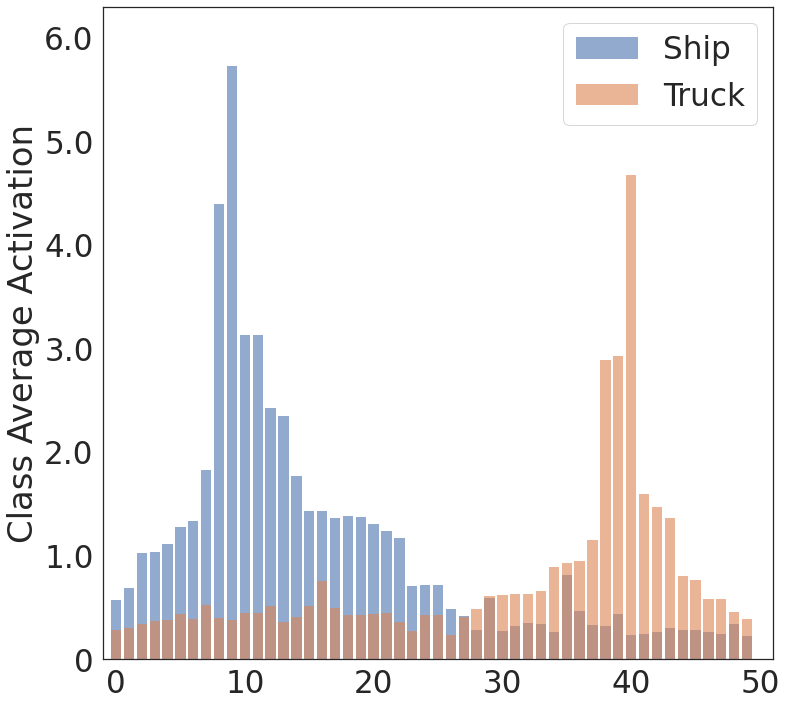}
         \captionsetup{justification=centering}
         \caption{Visual Words Activation\\ of Dissimilar Categories}
         \label{fig:unshared}
     \end{subfigure}
     \caption{Visual words exploration. (a) shows category similarity matrix, where each sample in \textit{STL10} is presented as a histogram over learned visual words and each category is the average of the samples in that category. (b)\&(c) demonstrate the visual word distributions on two similar categories and two dissimilar categories, where the x-axis denotes the 50 learned visual words.}
  \label{fig:word}\vspace{-3mm}
\end{figure}

\begin{figure}[t]
  \centering
  \includegraphics[height=6.5cm, width=\textwidth]{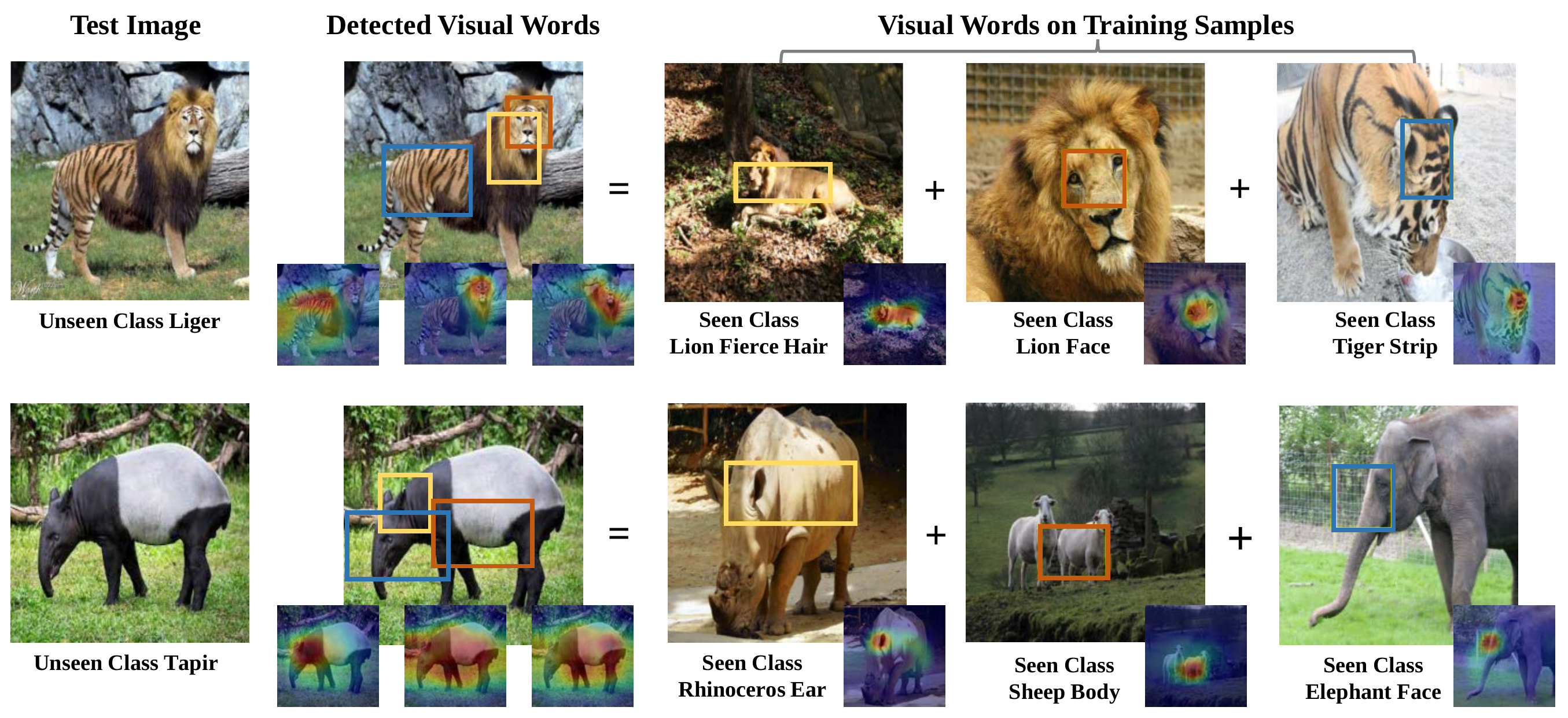}
  \caption{Illustrative examples of our learnable visual words trained on \textit{AWA2} for interpreting image samples of Liger and Tapir. Their categories are unobserved in the training set.}
  \label{fig:unseen}\vspace{-3mm}
\end{figure}

\subsection{In-depth Analyses}
In this section, we explore some aspects of our proposed model in more detail. Specifically, we examine how cross-category visual words help understand the relationship of categories. And we extend the application of our method to unseen categories during model training to elucidate the effectiveness of learnable visual words.

\textbf{Visual Words Exploration}. We first explore the learnable visual words by inspecting the interrelation between different classes. As we discussed in Section \ref{Prelim}, the output of visual words learning module $g \circ f(x_i)$ can be viewed as an activation vector representing how strong each visual word is associated with training image $x_i$. By averaging the vectors of all training samples in the category $c$, we acquire a new vector representation of the category $c$ defined by its similarity with the $M$ learned visual words. We calculate this activation vector for ten classes in \textit{STL10} dataset, and plot their pairwise correlations as a heatmap in Figure~\ref{fig:word}(a). We choose two pairs of classes, one similar and one dissimilar, and plot each pair's visual words based activation vectors together in Figure~\ref{fig:word}(b)\&(c). As demonstrated, the similar pair of {``Bird''} and {``Airplane''} has relatively high activations on at least ten visual words, which suggests these shared visual words' presence in both categories, while those unshared visual words differentiate one class from another. On the other hand, the dissimilar pair {``Truck''} and {``Ship''} barely agree on any visual words.

\textbf{Interpreting Samples From Unseen Categories}. In light of the visual words across different categories, we extend our method to those categories that potentially share semantics with existing training categories, but are unseen by the base model. In Figure~\ref{fig:unseen}, we choose {``Liger''} and {``Tapir''} as the unseen categories and test them on our model trained on \textit{AWA2} dataset. Ligers are zoo-bred hybrid offsprings of a male lion and a female tiger, which possess features of both parents. Tapirs are rare creatures that are often confused with pigs or anteaters, but their closest living relatives are actually rhinoceros and horses. As expected, our model successfully uncovers several shared semantics between the unseen classes with classes that exist in the training data. The face, fierce hair and strip pattern on the body of this liger are tracked back to the learned visual words from seen lion and tiger categories in the first row of Figure~\ref{fig:unseen}. Although our model does not predict this unseen image as a liger, it does help us to better understand this image: an animal with the face and fierce hair of a lion, as well as a tiger's stripe on its body. Similarly, our model can interpret this tapir as a creature, which has the ear of a rhinoceros, the body of a sheep and the face of an elephant.


\section{Conclusion}

To sum up, we presented Learnable Visual Words (LVW), a novel approach that aims to answer the question of \textit{how} deep models make predictions by extracting cross-category visual words and aligning learned visual words with the base deep models' visual attention. LVW simplifies ProtoPNet\cite{chen2019looks} by relaxing its class-specific constraint. Its attention guided semantic alignment preserves high fidelity with the base deep model on the level of predictive ability and model attention. We conducted extensive experiments on six benchmarks to quantitatively our model with recognition accuracy and our proposed IoU coverage metric against three state-of-art prototype-based methods. The experimental results evinced the efficacy and generalization of our LVW method.

{
\small
\bibliographystyle{abbrv}
\bibliography{references}
}

\end{document}